\documentclass[sigconf]{acmart}

\usepackage{times}
\usepackage{soul}
\usepackage{url}
\usepackage{amsmath}
\usepackage{amsthm}
\usepackage{booktabs}
\usepackage{algorithm}
\usepackage{algorithmic}
\usepackage[switch]{lineno}
\usepackage{subcaption}

\AtBeginDocument{%
  \providecommand\BibTeX{{%
    Bib\TeX}}}

\copyrightyear{2023} 
\acmYear{2023} 
\setcopyright{acmlicensed}\acmConference[AIES '23]{AAAI/ACM Conference on AI, Ethics, and Society}{August 8--10, 2023}{Montréal, QC, Canada}
\acmBooktitle{AAAI/ACM Conference on AI, Ethics, and Society (AIES '23), August 8--10, 2023, Montréal, QC, Canada}
\acmPrice{15.00}
\acmDOI{10.1145/3600211.3604706}
\acmISBN{979-8-4007-0231-0/23/08}

\begin{document}

\title{REFRESH: Responsible and Efficient Feature Reselection guided by SHAP values}


\author{Shubham Sharma}
\affiliation{%
  \institution{J.P. Morgan AI Research}
  \country{USA}
}
\email{shubham.x2.sharma@jpmchase.com}

\author{Sanghamitra Dutta}
\affiliation{%
  \institution{University of Maryland, College Park}
  \country{USA}
}
\email{sanghamd@umd.edu}

\author{Emanuele Albini}
\affiliation{%
  \institution{J.P. Morgan AI Research}
  \country{UK}
}
\email{emanuele.albini@jpmorgan.com}

\author{Freddy Lecue}
\affiliation{%
  \institution{J.P. Morgan AI Research}
  \country{USA}
}
\email{freddy.lecue@jpmchase.com}

\author{Daniele Magazzeni}
\affiliation{%
  \institution{J.P. Morgan AI Research}
  \country{UK}
}
\email{daniele.magazzeni@jpmorgan.com}

\author{Manuela Veloso}
\affiliation{%
  \institution{J.P. Morgan AI Research}
   \country{USA}
}
\email{manuela.veloso@jpmchase.com}

\renewcommand{\shortauthors}{Sharma et al.}

\begin{abstract}
  Feature selection is a crucial step in building machine learning models. This process is often achieved with accuracy as an objective, and can be cumbersome and computationally expensive for large-scale datasets. Several additional model performance characteristics such as fairness and robustness are of importance for model development. As regulations are driving the need for more trustworthy models, deployed models need to be corrected for model characteristics associated with responsible artificial intelligence. When feature selection is done with respect to one model performance characteristic (eg. accuracy), feature selection with secondary model performance characteristics (eg. fairness and robustness) as objectives would require going through the computationally expensive selection process from scratch. In this paper, we introduce the problem of feature \emph{reselection}, so that features can be selected with respect to secondary model performance characteristics efficiently even after a feature selection process has been done with respect to a primary objective. To address this problem, we propose REFRESH, a method to reselect features so that additional constraints that are desirable towards model performance can be achieved without having to train several new models. REFRESH's underlying algorithm is a novel technique using SHAP values and correlation analysis that can approximate for the predictions of a model without having to train these models. Empirical evaluations on three datasets, including a large-scale loan defaulting dataset show that REFRESH can help find alternate models with better model characteristics efficiently. We also discuss the need for reselection and REFRESH based on regulation desiderata. 
\end{abstract}



\received{20 February 2007}
\received[revised]{12 March 2009}
\received[accepted]{5 June 2009}

\maketitle

\section{Introduction}

\begin{figure*}
    \centering
    \includegraphics[width=.95\textwidth]{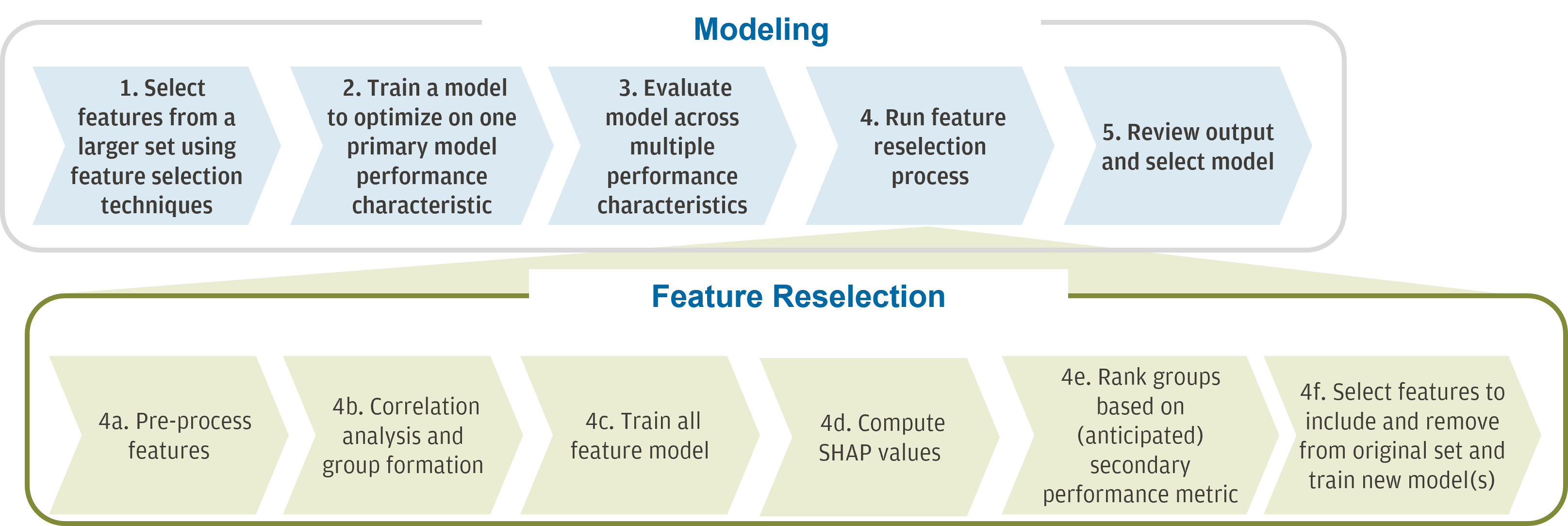}
    \caption{Standard model training and the framework for REFRESH. The top block shows the conventional steps to train a model (additional steps may also be used for model training, but we show the ones most relevant to the problem). This paper introduces the feature reselection process in step 4. The bottom block describes REFRESH}
    \label{fig:difference}
\end{figure*}

Machine learning models are increasingly being used in pivotal and sensitive industries such as finance~\cite{dixon2020machine,ahmad2018interpretable,babuta2018machine}, where big tabular datasets having millions of records across hundreds of dimensions (features) is common. Among a plethora of challenges, one question that model developers face is feature selection~\cite{kumar2014feature,bolon2015recent}. Feature selection aims to reduce the dimensionality of the data used to train the model while maintaining a model performance characteristic, which is often a measure of model accuracy. 

However, considerations beyond a specific measure of accuracy are imperative. Such model performance characteristics can include, but are not limited to pillars of responsible artificial intelligence \cite{dignum2019responsible,bhatt2020explainable}: fairness, explainability, and robustness. These characteristics towards building trustworthy models are essential to satisfy regulations \cite{carter2020regulation,madhavan2020toward,spiess2022machine}. When features are selected based on one primary model performance characteristic, such as accuracy, features that contribute towards secondary characteristics could have been dropped. This could occur in two ways: (a) features that make a secondary characteristic better were dropped, and (b) features that make a secondary characteristic worse were included.

When machine learning models have already been deployed with features selected based on a primary characteristic, a potential solution is to go back to the original model development process and select features with multi-objective characteristics to account for secondary characteristics. \cite{quinzan2022fast,dorleon2022feature}. However, feature selection in large-scale datasets is an expensive process \cite{Bommert2020a}. Furthermore, multiple objectives could be at odds with each other \cite{sharma2021fair,dutta2020there,zhang2019theoretically} and selecting features satisfying more than one objective still remains non-trivial. As research in responsible AI and regulatory requirements for machine learning models rapidly advance, new metrics are being developed to evaluate model performance, both within \cite{pagano2023bias,guo2023comprehensive} and beyond \cite{song2020systematic} the secondary characteristics discussed above. 
As the research community further investigates and devises these metrics, starting model development from scratch for existing models to optimize on new metrics is extremely expensive.

Hence, we introduce the process of \emph{feature reselection}. Feature reselection aims to select features to improve on secondary model performance characteristics (characteristics that become important to consider after a model is already developed) while maintaining similar performance with respect to a primary characteristic based on which features were already selected. Hence, reselection tries to find feature subsets that: a) include features that improve the secondary characteristic compared to the secondary characteristic of the model trained using features selected, b) do not include features that are detrimental to the secondary characteristic, and c) do not differ significantly from the feature subset that was used to train the model to optimize on a primary characteristic to maintain performance.

Reselection is not just 
useful for a model developer to save on time and effort when a model has already been deployed, but can be an extremely valuable tool for model monitoring. Specifically, the reselection process is agnostic to model metrics and can be run by a third-party monitoring the model. The process can help get insight into features that should or should not have been considered in the modeling pipeline, with respect to sensitive factors. Such information may not be available to a modeler. For example, in characteristics such as fairness, regulations require that sensitive attribute information and strong proxies to sensitive information are not available to modelers as features \cite{xiang2019legal,spiess2022machine}. Feature selection with fairness as a constraint becomes a much harder problem in the absence of the protected attribute. 
In these cases, the reselection process can then be done by a third-party that stores the sensitive information \cite{veale2017fairer}, to then suggest feature changes to modelers that can enhance fairness (such features would be weak proxies to sensitive information and do not provide direct information about the protected attribute, in accordance with legal requirements \cite{spiess2022machine}).

To address the problem of feature reselection, this paper introduces REFRESH: Responsible and Efficient Feature Reselection guided by SHAP values. REFRESH is agnostic to the model type (only requires prediction probabilities of a model) and to the primary and secondary performance characteristics (only requires a score for any model characteristic). The framework for REFRESH is shown in Figure~\ref{fig:difference}. Key steps in conventional machine learning model development involve feature selection, training a model to optimize on a primary performance characteristic, and evaluating the model along this characteristic before deployment. 
However, when the model is evaluated along a secondary characteristics, the same model may perform poorly. This is where the process of feature reselection is introduced, rather than re-computing models from scratch. 

Originally, to reselect features, a modeler would train new models on new feature subsets, across various trials of different feature subsets. This process can be very expensive with a large number of features. Additionally, it is hard to accomplish if the secondary characteristic computation requires sensitive information, since this is not available to a model developer. 
Hence, we introduce an efficient way to find alternate feature sets, without having to train a large set of new models. The feature reselection steps are shown in the bottom block of Figure \ref{fig:difference} and the steps are as follows: a) pre-process the set of all features; b) perform correlation analysis to create disjoint sets of groups of features, where groups are formed based on correlation between features (to be used in step e); c) train a model with all features; d) compute SHAP values (\cite{lundberg2017unified}) for each feature used to train the all feature model; e) use the SHAP values to approximate for model outcomes of models that would have been trained by removing each group and then rank each group of features formed in step b) based on anticipated effect of features on a secondary model performance characteristic; and f) select features to remove from the set of features selected by the modeler that have the most negative effect on the secondary characteristic and select features to include from the set of features that were not selected by the modeler that have the most positive effect on the secondary characteristic. Finally, train new models using these sets and provide alternate models.

The spine of REFRESH lies in using correlation based grouping of features and utilizing the additive property of SHAP values based feature attributions. SHAP \cite{lundberg2017unified} is a popular feature attribution technique \cite{bhatt2020explainable} and follows the additive property: the feature attributions sum to the model prediction for a given input. We show that combining this property of SHAP values with correlation analysis on groups of features provides a reasonable approximation to model outcomes of models trained in the absence of a group of features, without having to actually train these models. This significantly speeds up the ability to search for alternate models that can improve performance.

We show that REFRESH can help "refresh" a model to accommodate secondary characteristics i.e. find alternate models along multiple secondary characteristics, by experimentation on three datasets, including a large-scale loan defaulting dataset. The discussion section provides further insight into why reselection is needed, limitations of REFRESH, and the applicability of REFRESH based on regulations \cite{spiess2022machine}. The key contributions of this paper are\footnote{This works goal is not to provide models with optimal characteristics. Instead, the paper aims to introduce the research problem of feature reselection and provide a possible method to efficiently do this reselection. REFRESH can help find models that can perform better along multiple characteristics, but there are no guarantees on optimality. This is discussed further in experiments.}:
\begin{itemize}
    \item Introducing and motivating the research problem of feature reselection for incrementally improving secondary model characteristics;
    \item A novel approximation to model outcomes that uses grouping of features based on correlations, and SHAP values;
    \item REFRESH: an efficient method to reselect features that leverages this approximation.
\end{itemize}

\section{Related Work}

While the concept of reselection is new (to the best of our knowledge), this section points to resources for related work in the fields of feature selection, responsible AI, and within responsible AI, SHAP values. 

\vspace{0.1cm}
\noindent{\bf Features selection} has been a well studied problem in the machine learning literature. \cite{li2017feature,miao2016survey} cover the most popular feature selection methods, with an emphasis on selection based on accuracy as a performance objective. 

\vspace{0.1cm}
\noindent{\bf Responsible AI} includes fairness, adversarial robustness, explainability, and privacy of machine learning models \cite{schiff2020principles}. Models are considered more interpretable if less features are used to train the model \cite{poursabzi2021manipulating}. Feature selection based on fairness considerations \cite{grgic2016case,grgic2018beyond,quinzan2022fast,GalhotraSSV22,dutta2021fairness} is a growing field of research. Recently, \cite{dorleon2022feature} suggest a feature selection technique with both fairness and accuracy considerations. The method requires access to protected attributes, which are often not available. REFRESH only requires a fairness score, which can be provided using privacy-preserving methods \cite{chang2021privacy,fioretto2022differential}. \cite{yan2021cifs} propose a feature-importance-based improvement to adversarial robustness for CNN's. \cite{bakker2019fairness} discuss a method for fairness-based feature selection under budget constraints. Features selection with considerations on adversarial robustness for models trained using tabular datasets remains an unexplored problem. 

\vspace{0.1cm}
\noindent{\bf SHAP} (SHapley Additive exPlanations) \cite{lundberg2017unified}, a game theoretic approach to explain the output of any machine learning model, is a widely used technique in explainability of machine learning models. It is used to provide the feature importance for every feature used to train a model with extensions for fairness~\cite{begley2020explainability}. \cite{cohen2005feature} propose a method for feature selection using SHAP values. \cite{Fryer2021a} provide a detailed analysis on using SHAP values for feature selection. \cite{dong2021multi} use SHAP values of features for feature selection by using these values in a multi-objective optimization problem. \cite{marcilio2020explanations} show that SHAP values based selection performs better than three other feature selection techniques. 

\section{REFRESH: Theory and Method}

This section presents the theory, the core REFRESH method, and additional constraints that can be important for the feature reselection problem. 

\noindent{\bf Setup:} Consider a dataset with $\mathbf{N}$ features. The set of all features is $\mathbf{S_{N}}$. Let a feature selection method select a set of features to train a model for binary classification. The selected set of features is called the baseline set $\mathbf{S_{b}}$. Let the remaining feature set be the candidate set $\mathbf{S_{c}}$. Then:
\begin{equation}
    \mathbf{S_{b}} \cup \mathbf{S_{c}} =  \mathbf{S_{N}}
\end{equation}
\begin{equation}
    \mathbf{S_{b}} \cap \mathbf{S_{c}} = \{\}
\end{equation}

\vspace{0.1cm}
\noindent{\bf Correlation Analysis:} Step 4a in Figure \ref{fig:difference} requires pre-processing $\mathbf{S_{N}}$. Then, construct a graph of pairwise correlations between features and use a clustering algorithm to get groups of similar features (step 4b in Figure \ref{fig:difference}). Let $\mathbf{G_{i}}$ represent the $i^{th}$ group. If $k$ groups are formed then:

\begin{equation}
    \mathbf{G_{1}} \cup \mathbf{G_{2}}.. \cup\mathbf{G_{i}}.. \cup \mathbf{G_{k}} =  \mathbf{S_{N}}
\label{eq:groups}
\end{equation}
\begin{equation}
    \mathbf{G_{i}} \cap \mathbf{G_{j}} = \{\} \; \forall 1\leq i \leq k, 1 \leq j \leq k, i\neq j
\end{equation}

Consider a machine learning model $f$ trained on the all feature set $\mathbf{S_{N}}$ (step 4b in Figure \ref{fig:difference}) such that the prediction probability ${\mathbf{y}}$ for a given input instance $\mathbf{x}$ is:

\begin{equation}
    {\mathbf{y}}_{\mathbf{S_{N}}} = f(\mathbf{x_{S_{N}}})
\end{equation}

\vspace{0.1cm}
\noindent{\bf SHAP Values Computation:}
Compute the SHAP values of every feature in $\mathbf{S_{N}}$ for model $f(\mathbf{S_{N}})$. Let the SHAP value for feature $a$ for an input instance $\mathbf{x}$ be $\phi_{a}^{\mathbf{x}}$. SHAP values follow the additive \footnote{a.k.a., \emph{efficiency} in game theory \cite{Shapley1951}.} property \cite{lundberg2017unified}:
\begin{equation}
    {\mathbf{y}}_{\mathbf{S_{N}}} = \sum_{p=1}^{N} \phi_{p}^{\mathbf{x}}
\label{eq:shapadd}
\end{equation}

In other words, SHAP values can be understood as a (local) linear model approximating the contribution of each feature when included \cite{lundberg2017unified}. For a given input, the sum of these contributions equals the prediction probability of the model output for this input. Therefore, we could calculate (anticipatedly) the outcome of a model when a feature $a$ is absent as:
\begin{equation}
    {\mathbf{y}}_{\mathbf{S_{N}}\backslash a} = {\mathbf{y}}_{\mathbf{S_{N}}}- \phi_{a}^{\mathbf{x}}
\label{eq:shapdiff}
\end{equation}

\begin{figure*}
    \centering
    \includegraphics[width=.6\textwidth]{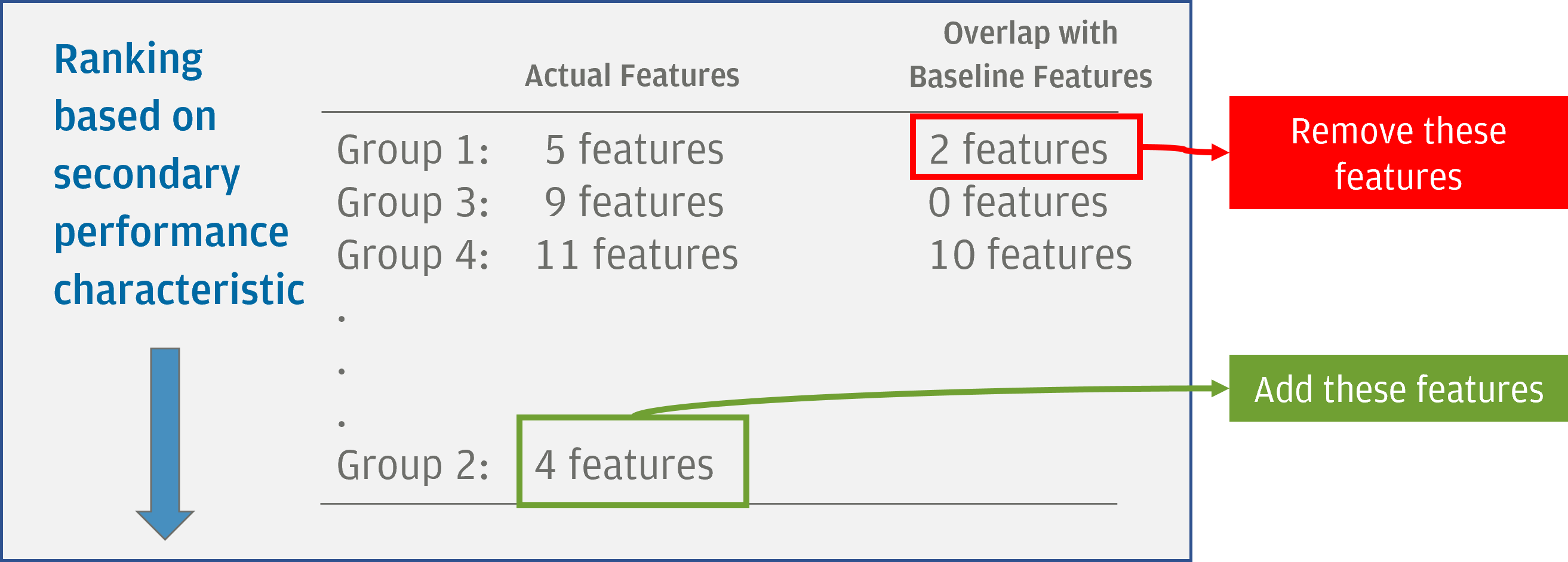}
    \caption{Ranking of groups based on secondary performance characteristic and the reselection process using this ranking.}
    \label{fig:ranking}
\end{figure*}


However, this calculation will not be accurate and outcomes can significantly differ from true model outcomes i.e. when feature $a$ is not used to train the model \cite{Chen2020,Fryer2021a,Kumar2021,dutta2022quantifying}. In fact, (interventional) SHAP simulates the removal of features by marginalising over their marginal distributions and not by re-training a new model without such features \cite{lundberg2017unified}. 
A simple example of why this occurs is as follows: if feature $a$ is perfectly correlated with $b$ and a model is trained using just $a$ and $b$, it may happen that the model used only feature $a$, therefore $b$ will have a SHAP value of 0 and the SHAP value of $a$ will be equal to the model prediction probability.
However, if feature $a$ is removed and a model is trained with just feature $b$, the outcomes would be the same as the first model, but the model outcomes calculated using Equation \ref{eq:shapdiff} would be $0$. 

\vspace{0.1cm}
\noindent{\bf SHAP Values based Approximation:}
REFRESH posits anticipating model outcomes based on the removal of a group of features, where features are grouped based on correlations. Combining Equations \ref{eq:groups} and \ref{eq:shapdiff}, approximate that:

\begin{equation}
    {\mathbf{y}}_{\mathbf{S_{N}}\backslash \mathbf{G_{i}}} \approx {\mathbf{y}}_{\mathbf{S_{N}}} - \sum_{p=l}^{m} \phi_{\mathbf{p}}^{\mathbf{x}}
\label{eq:approx}
\end{equation}
where,
\begin{equation}
     \mathbf{G_{i}} = \mathbf{S_{l,...,m}}
\end{equation}

Equation \ref{eq:approx} gives a better approximation when compared to directly using SHAP values. It is used to anticipate model outcomes without having to retrain new models (we show that this approximation is better empirically). Specifically, this enables REFRESH to anticipate (approximately) the outcome of a model when a group is absent from model training.

\vspace{0.1cm}
\noindent{\bf Feature Removal and Inclusion:}
These anticipated model outcomes can then be used to calculate anticipated secondary performance characteristics. For each $\mathbf{G_{i}}$, use the anticipated model outcomes and calculate an anticipated score of the secondary characteristic for each anticipated model, where each model corresponds to a model trained with the feature subset ${\mathbf{S_{N}}\backslash \mathbf{G_{i}}}$. Note that the score computation can be done by a third-party, thereby ensuring that sensitive information is not revealed to a model developer \cite{spiess2022machine} for secondary characteristics like fairness. The groups are then ordered in decreasing order of scores.

Figure \ref{fig:ranking} shows a toy example with feature groups that are ranked based on an anticipated secondary performance characteristic. Group 1 is ranked highest, which means that the anticipated (secondary characteristic) performance of the model when Group 1 was excluded from training was highest. This means that features from Group 1 are anticipated to be the most detrimental to the secondary characteristic. Hence, starting from the baseline set (to maintain the performance based on the primary characteristic) we would want to select a model with the feature subset: 
\begin{equation}
   \mathbf{S_{reselected}} =  \mathbf{S_{b}} \backslash \mathbf{G_{1}}
\label{eq:remove}
\end{equation}

Furthermore, Group 2 is ranked the lowest, which means that a model trained by removing Group 2 has an anticipated secondary performance characteristic which is lower. This means that features from Group 2 could contribute to a better secondary performance characteristic, and hence we include features from this group. Hence, we would want to select a model with feature subset: 

\begin{equation}
    \mathbf{S_{reselected}} =  (\mathbf{G_{2}} \backslash (\mathbf{G_{2}} \cap \mathbf{S_{b}})  )
\label{eq:include}
\end{equation}

Generalizing Equations \ref{eq:remove} and \ref{eq:include}, the feature subset to train a model after removing group $G_{r}$ and including feature $G_{i}$ is:

\begin{equation}
    \mathbf{S_{reselected}} = (\mathbf{S_{b}} \backslash \mathbf{G_{r}}) \cup (\mathbf{G_{i}} \backslash (\mathbf{G_{i}} \cap \mathbf{S_{b}})  )
\end{equation}

This process of removal and inclusion can be continued for more groups to generate new feature subsets that can be used to train alternate models. We discuss the choice of number of groups that should be considered for inclusion or removal in the experiments section.

\subsection{Additional Constraints}

In the feature reselection process, features are being added and removed with the objective of improving the secondary performance characteristic while maintaining the primary performance characteristic. However, involving a human-in-the-loop may ensure that features are not erroneously included or removed. Examples of errors are:
\begin{itemize}
    \item Features that are important for a classification task based on human judgement, and that maybe obviously important for the primary characteristic, are removed. This can especially occur when the primary and secondary characteristic are inversely related for the data and model under consideration. These features are important to explain the model prediction \cite{spiess2022machine}. For example, address is removed in a housing price prediction problem (because it could serve as a proxy for race) when a modeler thinks this feature is most important. Let these features be $\mathbf{S_{RE}}$ (Where RE means Removal Error).
    \item Features that should not be included based on human judgement and were removed as a part of feature selection are now included in the reselected set. For example, a feature with a lot of noisy values from the data collection process was removed with human insight, but is now included because it erroneously contributes to the secondary characteristic. Let these features be $\mathbf{S_{IE}}$ (Where IE means Inclusion Error).
\end{itemize}

REFRESH can easily incorporate these additional constraints that can be provided by modelers, so that erroneous features are not included or removed. The final reselected set is:

\begin{equation}
    \mathbf{S_{reselected}}  \nonumber= \\
    ((\mathbf{S_{b}} \backslash \mathbf{G_{r}}) \cup \mathbf{S_{RE}}) \cup ((\mathbf{G_{i}} \backslash (\mathbf{G_{i}} \cap \mathbf{S_{b}})) \backslash \mathbf{S_{IE}}
\end{equation}

\section{Experiments}

This section presents the context of the experimentation, results on applying REFRESH and validation experiments on the SHAP values based approximation.

\subsection{Context and Setup}

\noindent{\bf Data:} Experiments are performed on three datasets: COMPAS~\cite{ProPublica}, HMDA~\cite{HMDA}, and the large-scale home credit default risk dataset~\cite{HCDR}. We have used existing work to pre-process datasets and select baseline features, and refer to those works here. Additionally, we ensure that protected attributes (race, gender, age) are removed for model training, in accordance with legal requirements for model development. Details on the datasets and models used for them can also be found through these references. For the home credit default risk dataset, information and details on pre-processing can be found in \cite{HCRDproject}. For the COMPAS dataset, we use methods as in \cite{grgic2016case}. For the HMDA dataset, we use the same pre-processing and baseline set as in \cite{sharma2022feamoe}. This section focuses on experiments for the home credit default risk dataset since it has a large set of features, but experiments to validate REFRESH using the other two datasets are also provided (The COMPAS dataset is particularly useful for a qualitative validation of feature reselection by comparing to feature selection used in \cite{grgic2016case}). 

HomeCredit is a company that provides installment lending to people with poor credit history. In 2017, they made anonymized data available on Kaggle which includes individual demographics and loan outcomes. The raw data consists of millions of records and a total of 649 features. We pre-process the data similar to~\cite{HCDR}, so the final number of observations considered are 307,511 and the total number of features are 466.

\vspace{0.1cm}
\noindent{\bf Objective:}
REFRESH is model performance characteristic agnostic, and only requires a score for any model characteristic so that models can be ranked based on this secondary characteristic. The goal of the experiments is not to show models with optimal performance; rather, we show that we can find multiple alternate models showing varied model performances, including better performance along the secondary characteristic using REFRESH, and this is much faster than having to use brute force based search for reselection. This is in accordance with the aim to find less discriminatory alternatives \cite{spiess2022machine}, when the secondary characteristic is fairness. Additionally, we show that the approximation using SHAP values that is proposed in this paper (Equation \ref{eq:approx}) performs better than using just SHAP values (Equation \ref{eq:shapdiff}). 

\vspace{0.1cm}
\noindent{\bf Primary Characteristic:}
To show the ability of REFRESH to suggest alternative models, we consider the model AUC to be the primary model performance characteristic.

\vspace{0.1cm}
\noindent{\bf Secondary Characteristics:}
Experiments are performed for two different secondary characteristics (evaluated independently): fairness and adversarial robustness. These are just illustrative measures, and other model performance characteristics can also be considered. For fairness, the secondary characteristic considered is demographic parity. Statistical parity difference is used to measure demographic parity \cite{bellamy2019ai}. Given a model trained on a dataset with a protected attribute $A$ having two groups $a$ and $b$, where $a$ is the sensitive group and $Y$ is predicted output (thresholded prediction probability), the statistical parity difference is defined as:

\begin{equation}
    SPD = P(\mathbf{Y}=1|A=a) - P(\mathbf{Y}=1|A=b)
    \label{eq:SPD}
\end{equation}

For robustness, we consider using the notion of the distance to the boundary in the model output space, similar to \cite{sharma2021fair}. Specifically, if a point is closer to the decision boundary, the point is less robust (vulnerable to perturbations), and correspondingly, the prediction probability ${\mathbf{y}}$ is closer to the decision threshold $\delta$ set for binary classification. For any model, this can be calculated as:

\begin{equation}
    ROB = |\delta - {\mathbf{y}}|
    \label{eq:ROB}
\end{equation}

\vspace{0.1cm}
\noindent{\bf Experimentation Setup:}
Experiments for the home equity credit risk dataset are performed as follows: first, features are selected with the primary performance characteristic of AUC. Similar to~\cite{HCRDproject}, features are pre-processed. Sensitive attributes are removed for model training and are only used to compute the fairness score. Then, an XGBoost model is trained on all the features left after pre-processing, and the most important features based on feature importance scores are selected. A model is then trained with these features ($\mathbf{S_b}$), and this is the baseline model. For the home credit default risk dataset, 184 features are selected as the baseline feature set.

\vspace{0.1cm}
\noindent{\bf Correlation Analysis:}
Simultaneously, all features after pre-processing are grouped based on correlation. This is done by using the popular Louvain method for community detection~\cite{blondel2008fast,macmahon2013community}. The method is a greedy optimization method that runs in time $O(n\cdot \log n)$ where $n$ is the number of nodes in the network. The correlation of features defines whether a feature belongs to a community, and a correlation threshold is passed to define what constitutes a high correlation. For the home credit default risk dataset, this threshold is set to 0.7. Experiments on varying this threshold are also provided.

\vspace{0.1cm}
\noindent{\bf Feature Reselection:}
An XGBoost model is trained using all features and SHAP values are computed for this model for every feature. For each group of features, the anticipated model outcome (prediction probabilities)) for a model that would be trained by removing this group is calculated using Equation \ref{eq:approx}.

Then, the secondary performance characteristic is calculated using Equations \ref{eq:SPD} (for fairness)) or Equation \ref{eq:ROB} (for robustness) using these anticipated outcomes. Groups are then ranked in descending order based on the value of the secondary performance measure. For each group, the intersection with the baseline set is also found. Then, features that intersect with the baseline set from the top groups are removed and features that do not intersect with the baseline set from the bottom sets are added. Two hyperparameters, one for maximum number of features that can be included and one for maximum number of features that can be removed, are used. 

Starting from the highest rank, for each group, all features are removed (up to the maximum removal limit; if the limit is reached for some features of a group, remove a random subset). Starting from the lowest ranked group, only a certain number of features are included per group, and then the next group is considered to include features, until the maximum limit of inclusions (this is also a hyperparameter called number of inclusions per group). We only include some but not all features because groups are formed based on correlations, and including too many features from a group will not significantly impact any change in the model.

\subsection{Results for the home credit default risk dataset}

\begin{figure*}[!ht]
    \centering
    \begin{subfigure}[b]{0.45\textwidth}
        \centering
        \includegraphics[width=.69\textwidth]{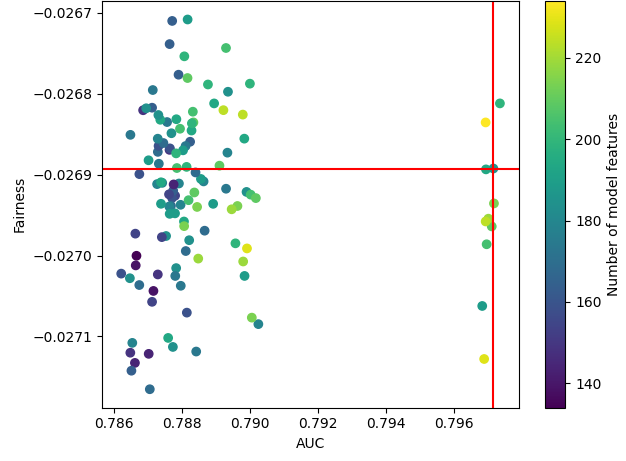}
        \caption{Fairness}
    \end{subfigure}
    \begin{subfigure}[b]{0.45\textwidth}
        \centering
        \includegraphics[width=.70\textwidth]{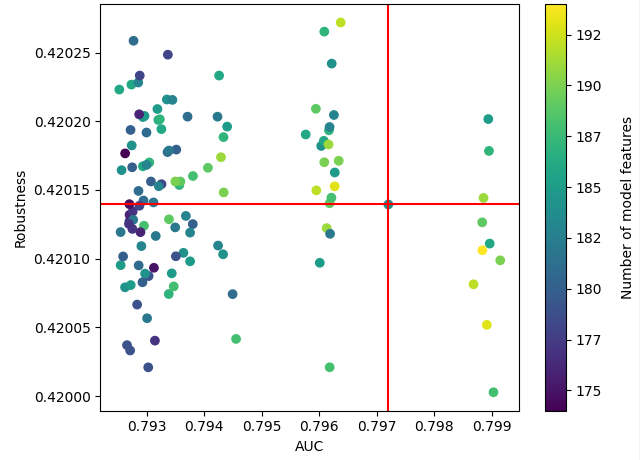}
        \caption{Robustness}
    \end{subfigure}
    \begin{subfigure}[b]{0.45\textwidth}
        \centering
        \includegraphics[width=.69\textwidth]{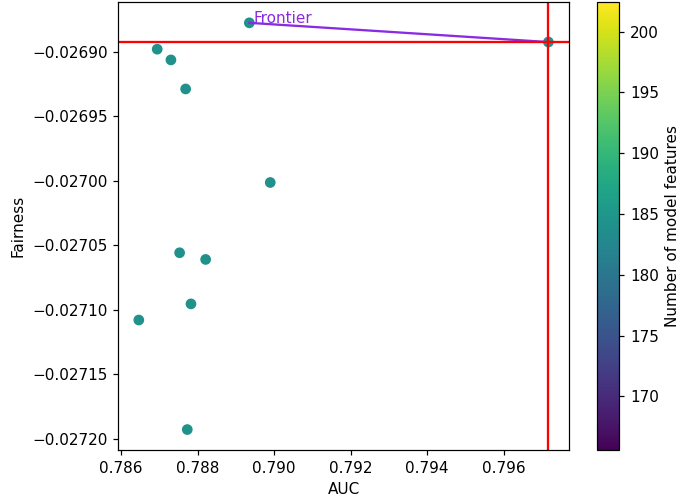}
        \caption{Fairness, $n(\mathbf{S_{reselected}}) = n(\mathbf{S_{b}})$ }
    \end{subfigure}
    \begin{subfigure}[b]{0.45\textwidth}
        \centering
        \includegraphics[width=.69\textwidth]{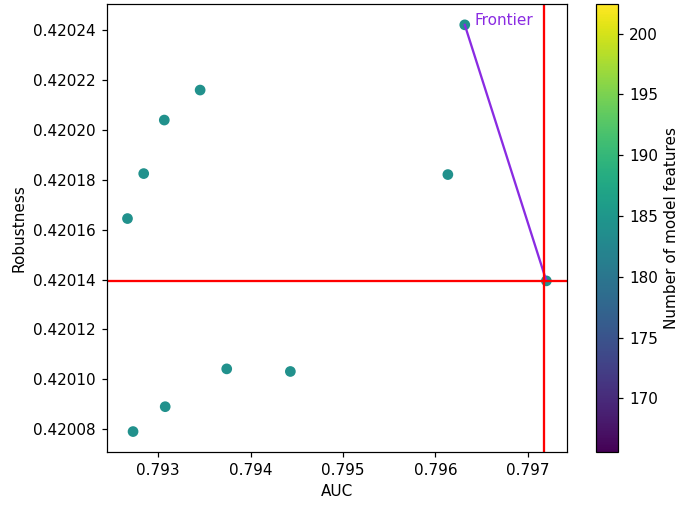}
        \caption{Robustness, $n(\mathbf{S_{reselected}}) = n(\mathbf{S_{b}})$}
    \end{subfigure}
    \caption{Alternate models found using REFRESH for two secondary performance characteristics: fairness ((a) and (c)) and robustness ((b) and (d)). Each point in the figure corresponds to a model trained using a different set of features. The intersection of the red lines is the baseline model. The reported metrics are the true measures and not anticipated values. The color of each point shows the number of features used to train the model. (b) and (d) show a subset of models from the fairness and robustness graphs (a) and (c) respectively, where each model has the same number of features as the baseline model.}
    \label{fig:Mainfigure}
\end{figure*}

\vspace{0.1cm}
\noindent{\bf On Fairness:}
Results for the fairness measure are shown in figure \ref{fig:Mainfigure}(a). All results are averaged over three runs. Each point on the graph is a model trained using a different subset of features (found using different inclusions and removals). The baseline model is marked by the intersection of the red lines. Starting with 5 features per group, and going up to a maximum of 50 features that can be included or removed, subsets are formed with combinations of inclusions and removals. Hence, one subset has 5 features included in the baseline, another has 5 features removed from the baseline, and a third would have a combination of these 5 included and the other 5 removed. This is done in increments of 5 features, until the maximum limit of inclusion and removals is reached (50). Hence, a total of 121 models is shown. The color of each model represents the number of features used to train the model.

The fairness of each model, in accordance with Equation \ref{eq:SPD} is plotted on the y-axis, and model AUC's are plotted on the x-axis. As we can see, several alternate models are found with varying degrees of fairness and AUC. It is interesting to note that while several models are found with an increase in fairness that also compromise on AUC (which is in accordance with expected trade-offs between fairness and accuracy \cite{dutta2020there}, there is one model with a larger set of features (compared to the baseline set) that has both a better fairness score and AUC. The increase in AUC is marginal, and within the threshold used to remove features in the original selection process. While it may appear that the increase in fairness is also marginal, the need to find less discriminatory alternatives still arises based on regulations \cite{spiess2022machine}, and the impact of a small increase on a dataset with millions of samples is more pronounced on individuals (more pronounced effects on fairness can be seen for the COMPAS dataset in experiments provided later). REFRESH provides a set of alternate models which can be chosen from, and the specific choice is dependent on the modeler or the regulator. It is key to note though that varied alternate models are found with just 121 more models being trained, as opposed to training a much larger set of models for hundreds of possible features.

While REFRESH does not guarantee optimality on models found with respect to any performance metric, it efficiently informs a modeler or regulator on the direction of the search space. Specifically, with being informed about which features can be added or removed to improve or reduce the secondary performance characteristic, far fewer models need to be trained, and the whole feature selection process does not need to be repeated. Further insight on alternate models can be gathered through an investigation such as the one shown in figure \ref{fig:Mainfigure}(c). The plot shows a subset of points from the fairness plot above, where every alternate model has the same exact number of features as the baseline model. Among these models, the frontier showcases two models, one that has the highest accuracy and the other that has the highest fairness. A modeler can decide which one to choose (based on which measure is more important to the application), while keeping the number of features to be similar to the baseline set (to maintain model complexity and explainability).

\begin{figure*}[!ht]
    \centering
    \begin{subfigure}[b]{0.4\textwidth}
        \centering
        \includegraphics[width=.8\textwidth]{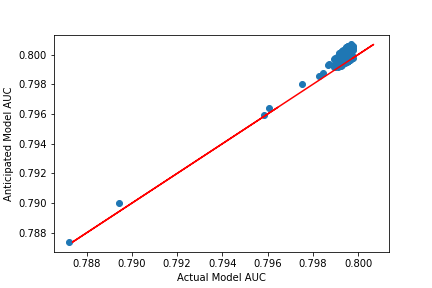}
        \caption{REFRESH (grouping features based on correlations)}
    \end{subfigure}
    \begin{subfigure}[b]{0.4\textwidth}
        \centering
        \includegraphics[width=.8\textwidth]{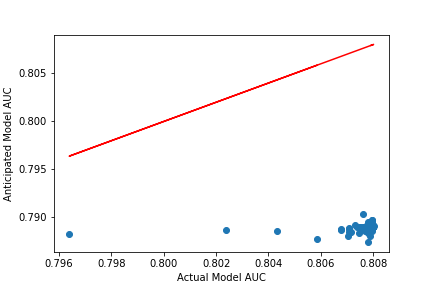}
        \caption{Without any grouping}
    \end{subfigure}
    \caption{Understanding the correlation grouping based SHAP approximation. Both graphs show the anticipated model AUC's against the actual model AUC. In (a), each point on the graph represents the anticipated versus actual AUC of a model trained with all features except all features from one group. In (b), each point on the graph represents the anticipated versus actual AUC of a model trained with all features except one feature (chosen at random) from one group. The red lines show the ideal plot (where anticipated AUC $=$ actual AUC).}
    \label{fig:Experimentapprox}
\end{figure*}

\vspace{0.1cm}
\noindent{\bf On Robustness:}
Similar results are shown for the robustness performance characteristic in figure \ref{fig:Mainfigure}(b) and (d). The hyperparameters for number of inclusions and removals (and limits) are the same as for the fairness plot. Better alternate models with respect to both robustness and AUC are found. However, it is clear that these models have more features than the baseline set. \ref{fig:Mainfigure}(d) shows models that have the same number of features as the original model. Results indicate that to maintain the same model complexity, a trade-off between AUC and robustness is required. A modeler or a stakeholder monitoring/regulating can choose which model suits the requirements for the specific task.

\subsection{SHAP values based approximation}

To check that the proposed SHAP values based approximation (Equation \ref{eq:approx}) of the model output performs better than using SHAP values without considering groups of features when groups are formed based on correlations, a comparison of two cases is done on the anticipated versus the actual model AUC, where: (a) the first case considers the AUC found for models trained (or anticipatedly trained) by removing an entire group from the all feature set, in accordance with Equation \ref{eq:approx}; (b) the second case considers model outcomes for models trained (or anticipatedly trained) by removing just one feature per group, in accordance with Equation \ref{eq:shapdiff}.

Results are shown in figure \ref{fig:Experimentapprox}. The red line indicates the ideal plot. The graph on the left shows anticipated outcomes against the actual outcomes when anticipated outcomes are found using the approximation used in REFRESH. Anticipated model AUC's are relatively close to those of ideal models, showing that the SHAP approximation using groups of features holds reasonably. On the other hand, when correlated features are not grouped together, the anticipated outcomes of the removal of individual features are incorrectly estimated by Equation \ref{eq:shapdiff}. The anticipated AUC is always less than the actual AUC. This happens because the anticipated outcome is based on the SHAP value of the feature to be removed. When the actual model is trained (with the removal), another feature belonging to the same group can take a higher SHAP value than what it had before (replacing the effect of the old feature). Hence, the true effect of removal is minimal, but seems more pronounced by using SHAP without grouping features to find anticipated outcomes.

\subsection{Additional Details and Experiments}

\subsubsection{Effect of correlation threshold on the SHAP approximation}

The SHAP approximation relies on forming groups of correlated features to find anticipated model outcomes when each group is removed. Hence, how close the anticipated outcomes are to true model outcomes depends on the correlation threshold for group formation. To study this, we find the difference in the anticipated and actual model AUC's for models formed by removing each of the groups. Fairness and robustness also depend on model outcomes, and a difference between true and anticipated values of these characteristics observe a similar effect to AUC, so these plots have been omitted. 

\begin{figure*}[!ht]
    \centering
        \centering
        \includegraphics[width=.4\textwidth]{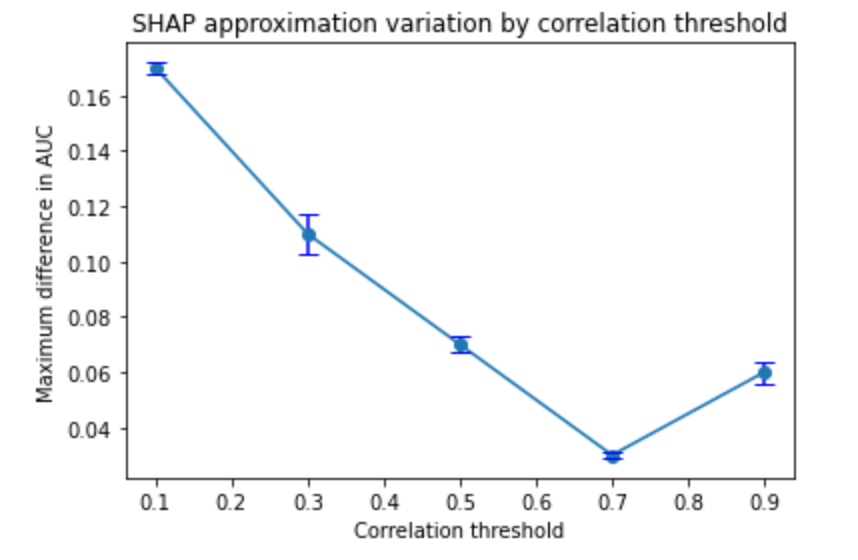}
    \caption{Analysing the proposed SHAP approximation (for the home credit default risk dataset) via plotting the difference (lower is better) in actual and anticipated AUC's versus the correlation threshold chosen to form groups.}
    \label{fig:SHAPapproxcorr}
\end{figure*}

Figure \ref{fig:SHAPapproxcorr} shows the maximum difference between anticipated and actual AUC's of models when different correlation thresholds are used to form groups. As we can see, very low correlation thresholds would cause the approximation to suffer much more than choosing a very high correlation threshold. However, a high correlation threshold would result in more groups being formed which would result in more calculations for anticipated outcomes and hence slower performance. Having a relatively higher (0.7) correlation threshold works the best, and this is observed across datasets.

\subsubsection{Model hyperparameters}

For the XGBoost model used for the home credit default risk dataset, 50 tree estimators are used with a maximum depth of 5. All other parameters are kept to default values for scikit learn's XGBoost model. For the COMPAS dataset, 2 estimators are used (since the dataset is very small) with a max depth of 3. For the HMDA dataset, 10 tree estimators are used with a maximum depth of 5.

As features are added or removed, hyperparameter tuning may have to be repeated. For the purposes of this paper, since we do not remove or add too many features, the hyperparameters are kept the same across different models since experiments showed that changing these had negligible impact on model performance. However, as more features are included or removed, tuning of parameters maybe required for optimal performance on alternate models.

\subsubsection{REFRESH hyperparameters}

Three hyperparameters are associated with REFRESH: the maximum number of inclusions, maximum number of removals, and number of inclusions per group. To show the difference in performance as these parameters vary, we report the AUC and fairness scores associated for models trained on the home credit default risk dataset for three different values associated with these parameters where the models are chosen such that they have the best secondary performance characteristic.

The results are shown in tables \ref{maxremovals}, \ref{inclusions} and \ref{inclusionpergroup}. As seen, having a small value for maximum removals or maximum inclusions yields sub-optimal performance on alternate models found. Having a very high value for these parameters does not help with the best model being found and would just increase the number of models being trained. For the number of inclusions per group, having a low value may result in some helpful features (with respect to the secondary characteristic) being neglected. Having a very high value does not help and just adds to the number of features, since inclusions are performed from groups of correlated features.

\begin{table*}[t!]
\caption{Varying the maximum number of removals hyperparameter for the home credit default risk dataset}
\centering
\begin{tabular}{cccc} 
\hline
    Performance &   Max removals =5 & Max removals = 50 & Max removals =100   \\
\hline
SPD   &  -0.2689  & -0.0267 & -0.0267\\
AUC     & 0.796 & 0.788 &  0.788\\
\hline
\end{tabular}
 \label{maxremovals}

 \end{table*}

 \begin{table*}[t!]
\caption{Varying the maximum number of inclusions hyperparameter for the home credit default risk dataset}
\centering
\begin{tabular}{cccc} 
\hline
    Performance &   Max inclusions =5 & Max inclusions = 50 & Max inclusions =75   \\
\hline
SPD   &  -0.272  & -0.0267 & -0.0267\\
AUC     & 0.797 & 0.788 &  0.788\\
\hline
\end{tabular}
 \label{inclusions}

 \end{table*}

 \begin{table*}[t!]
\caption{Varying the maximum number of inclusions per group for the home credit default risk dataset}
\centering
\begin{tabular}{cccc} 
\hline
    Performance &   Inclusion per group =1 & Inclusion per group = 3 & Inclusion per group=5   \\
\hline
SPD   &  -0.274  & -0.0267 & -0.02672\\
AUC     & 0.7955 & 0.788 &  0.788\\
\hline
\end{tabular}
 \label{inclusionpergroup}

\end{table*}

\subsubsection{Confidence intervals}

The average standard deviations for AUC, fairness and robustness measures are reported in table \ref{stdev}. The values are low, showing that results are consistent across runs.

\begin{table*}[t!]
\caption{Average standard deviation for different performance measures for the home credit default risk dataset when results are average across three runs}
\centering
\begin{tabular}{cc} 
\hline
Performance &   Standard deviation  \\
\hline
Accuracy    &  0.00823  \\
SPD      &  0.00194 \\
ROB      & 0.00042  \\
\hline
\end{tabular}
 \label{stdev}

\end{table*}

\subsubsection{Experiments on COMPAS and HMDA datasets}

\begin{figure*}[!ht]
    \centering
    \begin{subfigure}[b]{0.45\textwidth}
        \centering
        \includegraphics[width=.69\textwidth]{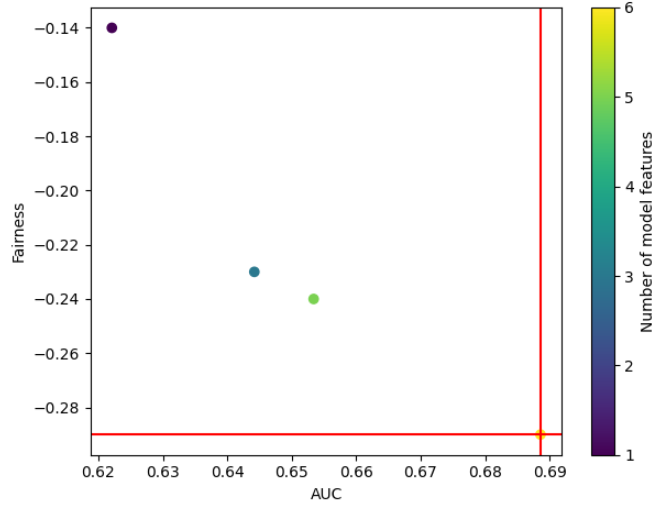}
        \caption{Fairness COMPAS}
    \end{subfigure}
    \begin{subfigure}[b]{0.45\textwidth}
        \centering
        \includegraphics[width=.70\textwidth]{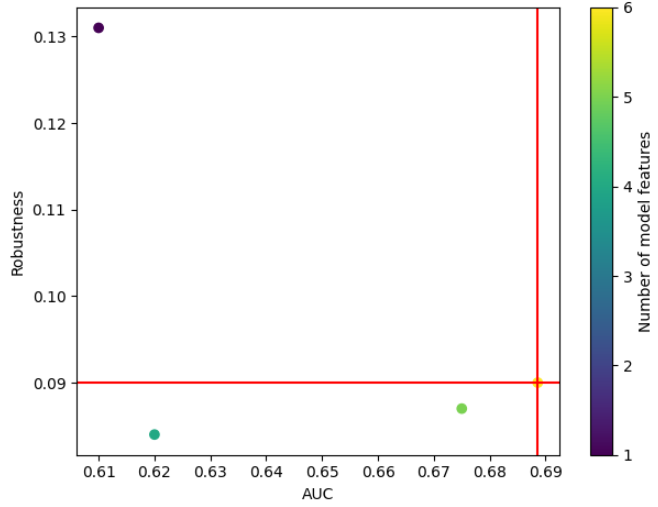}
        \caption{Robustness COMPAS}
    \end{subfigure}
    \begin{subfigure}[b]{0.45\textwidth}
        \centering
        \includegraphics[width=.69\textwidth]{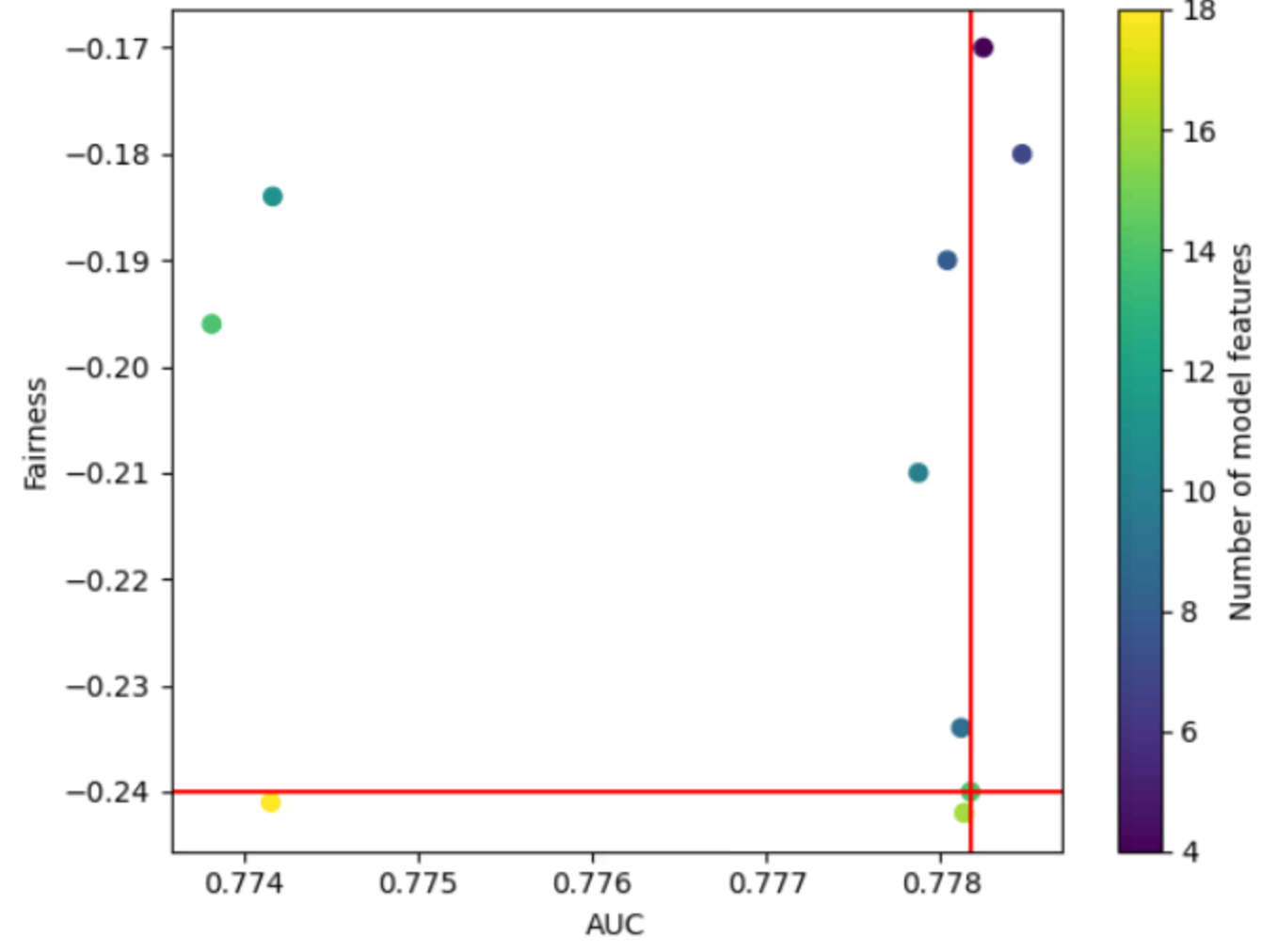}
        \caption{Fairness HMDA }
    \end{subfigure}
    \begin{subfigure}[b]{0.45\textwidth}
        \centering
        \includegraphics[width=.69\textwidth]{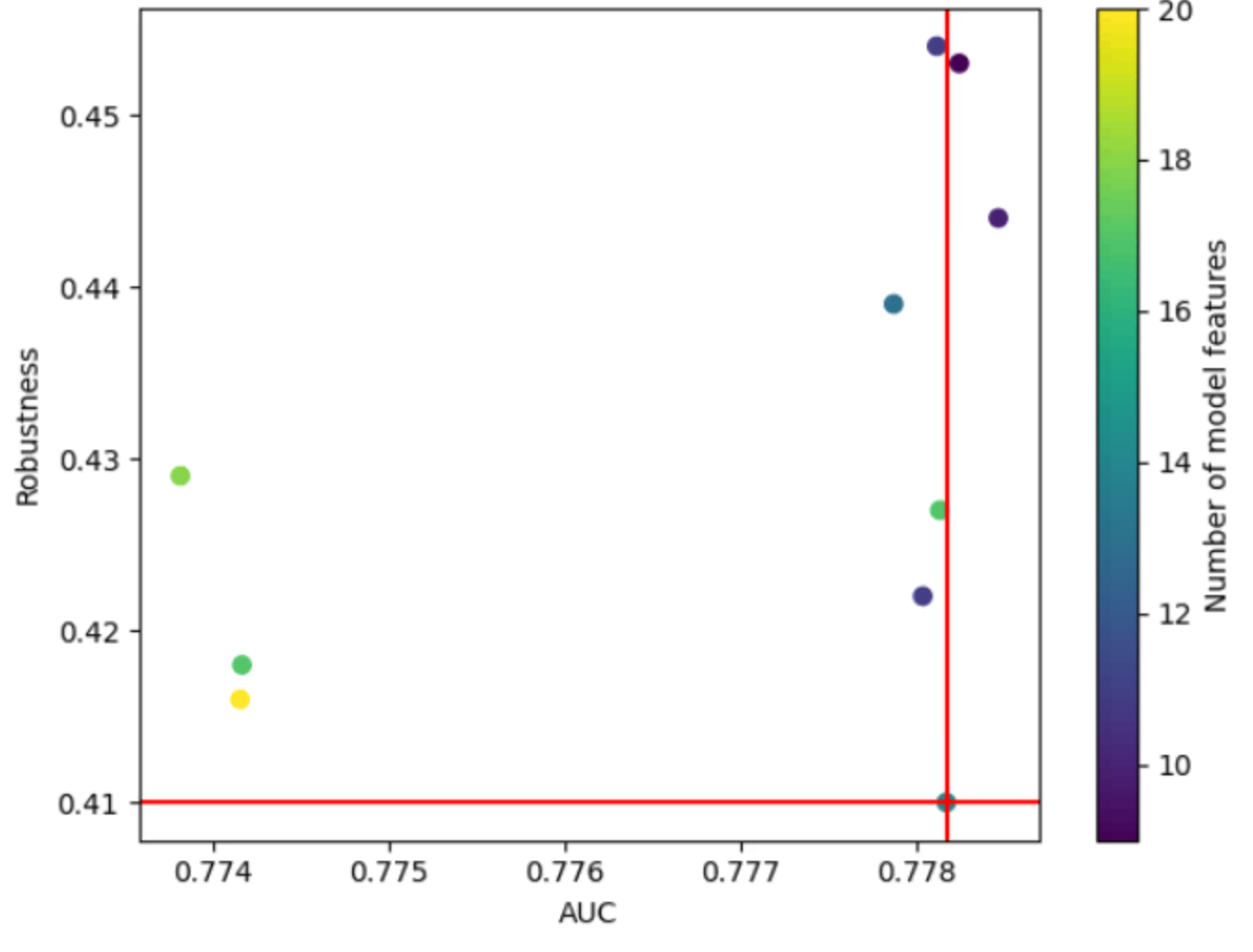}
        \caption{Robustness HMDA}
    \end{subfigure}
    \caption{Alternate models found using REFRESH for two secondary performance characteristics: fairness and robustness, and for two datasets: COMPAS (top) and HMDA (bottom). Each point in the figure corresponds to a model trained using a different set of features. The intersection of the red lines is the baseline model. The reported metrics are the true measures and not anticipated values.}
    \label{fig:Moreexp}
\end{figure*}

Results for alternate models for the two datasets are shown in Figure \ref{fig:Moreexp}. As can be seen, multiple alternate performance with different secondary characteristics can be found with just a few more models being trained.

Additionally, it is interesting to note that the best performance point in the COMPAS dataset with respect to fairness in figure \ref{fig:Moreexp} corresponds to just having one feature, which is the same feature found in \cite{grgic2016case} as the only feature being selected which is the most fair to judge recidivism (prior counts). Hence, REFRESH is able to automatically find feature sets that correspond to fairer features.

Finally, the COMPAS dataset has very few features, so finding more robust models is harder. This is shown in the robustness plot, where removing a few features resulted in robustness similar to the baseline model, but with a compromise on performance. However, the model is more robust when more features are removed. This analysis shows that eventually, the performance of REFRESH, just like any feature selection algorithm, is limited by the availability of features that can help with the secondary characteristic.

\subsubsection{Experiment on neural network}

\begin{figure*}[!ht]
    \centering
        \centering
        \includegraphics[width=.35\textwidth]{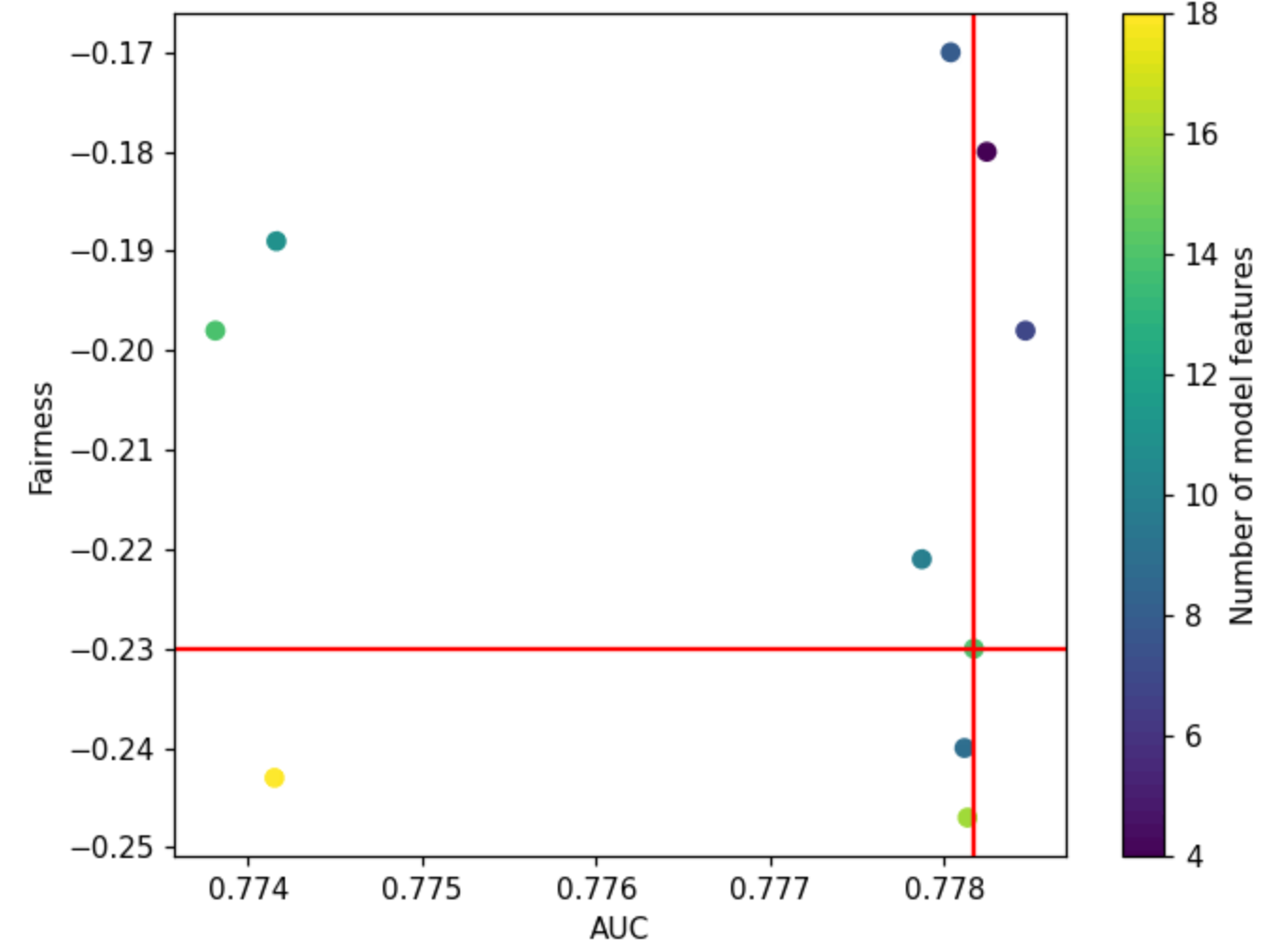}
    \caption{Results on fairness for a neural network trained using the HMDA dataset}
    \label{fig:Neuralnetwork}
\end{figure*}

To illustrate with an example that REFRESH is model agnostic, we perform an experiment on using a neural network with the HMDA dataset for the fairness characteristic. The neural network architecture is the same as in \cite{sharma2022feamoe}. The results are shown in figure \ref{fig:Neuralnetwork}. As we can see, the results are similar to the results in \ref{fig:Moreexp}. The key difference in implementation is in the use of KernelSHAP for the neural network as opposed to TreeSHAP for the XGBoost model. 

\section{Discussion}

This section is focused on discussions, including limitations, on the three novel components of this paper: feature reselection, REFRESH's methodology, and applicability of REFRESH based on regulations and insights from consumer lending \cite{spiess2022machine}.

\subsection{Feature Selection and Reselection}

Feature reselection is not introduced to replace responsible feature selection. Instead, it aims to provide an alternate efficient technique in cases where: a) models trained using a large set of features have already been deployed with selected features based on a primary characteristic and require re-evaluation for additional characteristics, b) new regulations require finding alternate models that improve based on secondary characteristics, and c) new research drives the need to evaluate models along different characteristics.

To achieve these objectives, REFRESH has been developed to aid model redevelopment. Since ranking of feature groups only depends on a score and not on the actual definition of the secondary characteristic, new secondary characteristic definitions can be readily incorporated to find alternate feature subsets. It does not replace the need for human insight on features that should be included or excluded, but is a tool that helps guide reselection based on desirable model characteristics. 

\subsection{Limitations of REFRESH}

A limitation of this method is that the accuracy of the REFRESH approximation depends on the structure and correlations of the data itself, and the ability to find groups of features based on correlations, such that these groups are disjoint. This may not always be possible, and the approximation may perform worse in cases where the disjoint groups of features cannot be formed easily. However, the method could still yield insights into features that help improve secondary performance characteristics. We note that resorting to alternatives such as Conditional and Causal SHAP \cite{Aas2021,Heskes2020,Frye2020} could mitigate this problem. However, on top of the technical challenges of estimating a causal graph of the features, doing so could result in features not used by the model having a non-zero importance, an issue certainly no less important in the feature reselection setting. Additionally, some other feature attribution techniques cannot be compared to because they do not follow the additive property, fundamental to use the approximation in Equation~\ref{eq:approx}.
Additionally, REFRESH hyperparameters may also require grid search, causing the efficiency to decrease to find alternate models. We leave the investigation of techniques to make REFRESH more efficient as future work.

\subsection{REFRESH and Regulatory considerations}

REFRESH is strongly motivated by findings from \cite{spiess2022machine}. Regulations require that model developers do not use sensitive information in any model development procedure for critical applications. Additionally, there is a growing need to find less discriminatory alternative models for such applications, such as in home lending.

REFRESH helps provide less discriminatory alternatives without requiring access to sensitive information (and just requiring a score for fairness which can be computed by a third-party). Furthermore, providing additional constraints to control features that cannot be added or removed are in accordance with insights for explainability in \cite{spiess2022machine}: features that can be explained by reason codes should be included. 

Privacy based secondary characteristics \cite{song2020systematic} can directly be used in the REFRESH framework to select features that can cause the most leakage of data information, and these can be removed. Formally analysing privacy considerations for REFRESH is left as future work.




\section{Conclusion and Future Work}

This paper introduces and motivates the problem of feature reselection. We then propose REFRESH: Responsible and Efficient Feature Reselection guided by SHAP values. REFRESH uses a combination of correlation analysis and the additive property of SHAP values to provide an approximation that can help find alternate models more accurately than directly using SHAP values. This can then be used to find models with improvements in secondary performance characteristics such as fairness and adversarial robustness. Experiments on three datasets, including a large-scale dataset in the finance domain, show that REFRESH can find several alternate models efficiently for multiple secondary performance characteristics.
There are a plethora of possibilities that can be explored as future work. New methods can be created to deal with feature reselection, such that they could be more optimal with respect to the secondary performance characteristic. We choose SHAP values because of their additive property, but other feature attribution techniques that follow this property can also be considered and compared to in the future. It would also be interesting to explore the ability to create groups of features that intersect inter-group so that the approximation is improved. Finally, the method can also be extended for experiments on additional secondary performance characteristics (eg. privacy).


\section*{Disclaimer}

This paper was prepared for informational purposes by
the Artificial Intelligence Research group of JPMorgan Chase \& Co\. and its affiliates (``JP Morgan''),
and is not a product of the Research Department of JP Morgan.
JP Morgan makes no representation and warranty whatsoever and disclaims all liability,
for the completeness, accuracy or reliability of the information contained herein.
This document is not intended as investment research or investment advice, or a recommendation,
offer or solicitation for the purchase or sale of any security, financial instrument, financial product or service,
or to be used in any way for evaluating the merits of participating in any transaction,
and shall not constitute a solicitation under any jurisdiction or to any person,
if such solicitation under such jurisdiction or to such person would be unlawful.

\bibliographystyle{ACM-Reference-Format}
\bibliography{ijcai23}

\end{document}